\documentclass{article}
\usepackage{ijcai26}

\usepackage{times}
\usepackage{soul}
\usepackage{url}
\usepackage[hidelinks]{hyperref}
\usepackage[utf8]{inputenc}
\usepackage[small]{caption}
\usepackage{graphicx}
\usepackage{amsmath}
\usepackage{amsthm}
\usepackage{booktabs}
\usepackage{amssymb}
\usepackage[ruled,vlined,linesnumbered]{algorithm2e}
\usepackage[font=small]{subcaption} 
\usepackage[table]{xcolor} 
\usepackage{booktabs} % Highly recommended by instructions 
\usepackage{multirow}
\usepackage{cleveref}               

\DeclareUnicodeCharacter{2212}{-}
\DeclareMathOperator*{\argmax}{arg\,max}
\DeclareMathOperator*{\argsort}{arg\,sort}

\newcommand{\TSS}{\emph{TSS}}

\title{INSIGHTS: Demonstration-Based Summaries of Time Series Predictors}

\author{
Bar Eini Porat $^1$\and
Rom Gutman $^1$\and
Uri Shalit $^2$\and
Ofra Amir $^1$
\\
\affiliations
$^1$ Technion Israel Institute of Technology \\
$^2$ Tel-Aviv University 
\emails
briany202@gmail.com,
}

\begin{document}

\maketitle

\begin{abstract}
{Explainability methods have progressed rapidly, but global explanations for time-series models remain underdeveloped, with most approaches focusing on local, instance-level attributions. We introduce INSIGHTS, a model-agnostic, user-centric approach for providing global explanations of time series models. Our approach prioritizes simplicity, efficiency, and transparency in its design, ensuring that stakeholders can readily adopt its outputs. While current methods focus on local explanations, INSIGHTS generates sample summaries that offer a comprehensive overview of model behavior. It balances the importance and diversity of time series samples to create informative subsets using utility functions that capture domain-specific aspects of time series behavior, such as exceeding domain norms. We evaluate INSIGHTS through experiments, interviews, and a user study. Our results indicate INSIGHTS effectively constructs comprehensive, diverse time series subsets, producing summaries manageable for individual evaluation. It is preferred by domain experts for its ability to provide a stable understanding of model behavior and the quality of the samples identified. Moreover, user study participants presented with INSIGHTS-based summaries exhibit an enhanced understanding of the model's overall behavior.}
\end{abstract}

\section{Introduction}\label{intro}

The growing availability of time series data increases the need for interpretability in time-series models. While interpretability for time series has lagged behind vision and NLP, interest in understanding time-series predictors is rising \cite{theissler2022explainable}, \cite{arsenault2025survey}. Theissler's review \shortcite{theissler2022explainable} highlights the strong progress in \emph{local} interpretability, including model-specific and model-agnostic methods that explain individual predictions. In many applications, however, experts also need \emph{global} explanations: a view of how a model behaves across regimes, typical cases, and critical events, not only behavior in a single instance \cite{jacobs2021designing}. Existing global methods for time series forecasting are still largely model- or modality-specific \cite{rojat2021explainable}. This is limiting in time-series prediction, where different model families often coexist (e.g deep learning or regression-based). As a result, despite the availability of local, model-agnostic explainers, there is (to our knowledge) no \emph{reliable} global, model-agnostic explanation approach tailored to time-series prediction—leaving unclear which cases should be inspected and which behaviors should be prioritized.

We propose \emph{INSIGHTS}, a model-agnostic algorithm for global explanations of time-series \emph{prediction} models (not restricted to classification). INSIGHTS produces time-series summaries: small sets of segments or events that highlight key model behaviors. The goal is to make local, model-agnostic explanations usable at scale by selecting a principled set of examples for inspection, rather than relying on arbitrary or convenient cases. A useful analogy is an executive summary: stakeholders do not review an entire dataset, but can audit a carefully chosen set of examples that reflects model performance in practice. Selecting these segments is challenging because “importance” is domain- and context-dependent and often driven by rare or unlabeled events. For example, in stock prediction, sharp surges or drops may matter more than routine oscillations; in the ICU, persistent increases in blood pressure may signal deterioration. INSIGHTS captures such preferences through user-defined utility functions for sequential data and combines them with a time-series–compatible notion of subset diversity, producing summaries that balance \emph{importance} and \emph{coverage} of distinct behaviors. We provide three utility functions capturing major aspects of time-series behavior, adapted from the utility scores of Eini Porat \shortcite{eini2024aiming} as default instantiations within INSIGHTS.

We evaluate INSIGHTS with an emphasis on whether summaries help people, not only whether they optimize proxy objectives. The evaluation includes: (i) an event-capturing experiment that measures recovery of important and diverse signal events, (ii) semi-structured interviews with clinical experts, and (iii) a user study on stock prediction where participants complete a predefined task to measure whether summaries improve participants' understanding of model behavior. We compare against three strong baseline approaches applicable to forecasting. Across settings, INSIGHTS produces summaries that match or outperform the strongest baseline on event capture while achieving higher diversity by both quantitative and qualitative measures. Compared to prototype-based alternatives, INSIGHTS is simple and transparent, and it is more computationally and memory efficient, making it a practical component for selecting meaningful subsets in larger interactive explanation systems.

\section{Background}\label{background}

In the XAI literature, \emph{global} explanations describe a model's overall behavior, while \emph{local} explanations describe behavior with respect to a specific input \cite{Doshi-Velez2017TowardsLearning}. For time series, local explanations are relatively mature: both time-series–specific methods and general tools such as SHAP and LIME \cite{ribeiro2016should,lundberg2017unified} can explain individual examples via point-wise attributions/attention, subsequence patterns (e.g., shapelets), or instance-based rationales (e.g., prototypes) \cite{rojat2021explainable}. In principle, if one could identify a small, representative set of time-series segments, these local tools could be applied to that set to yield a practical, end-to-end explanation.

The challenge is obtaining such a set in a \emph{global, model-agnostic} way for time-series \emph{prediction}. Global “example summary” approaches have been studied in classification \cite{Ribeiro2016WhyClassifier} and reinforcement learning \cite{Amir2018HIGHLIGHTS:People,Amir2019SummarizingStrategies,Lage2019TowardSummarization}, but these methods setting-specific (i.e., rely on class labels or RL value functions) and do not transfer directly to prediction settings. More broadly, model-agnostic subset selection methods such as ProtoDash and MMD-Critic \cite{gurumoorthy2019efficient,Kim2016ExamplesInterpretability} produce representative and diverse subsets in modalities like tabular and image data, yet applying them to time series is non-trivial: similarity, importance, and diversity depend on temporal context and can be dominated by rare events, regime changes, or transient anomalies. Unlike prototype methods (e.g., ProtoDash, MMD-Critic) that summarize the \emph{data distribution}, INSIGHTS summarizes \emph{model-relevant temporal behaviors} for prediction, emphasizing domain-defined, event-driven patterns that are often rare or asymmetric.

Model-agnostic methods for \emph{time-series prediction} remain scarce \cite{theissler2022explainable}; yet recent work extends global explanations for time series forecasting by aggregating feature attributions. TsSHAP \cite{raykar2023tsshap} learns a surrogate over interpretable features with local and global variants, where the global view aggregates absolute attributions across the dataset. However, for time series, dataset-level aggregation can blur event- and context-dependent effects, resulting in the reported lower fidelity for its global explanations \cite{raykar2023tsshap}.As TsSHAP provides global feature importance rather than example summaries, it is not directly comparable to our setting, though its local explanations could be applied to our selected samples in future work.

Finally, many time-series–specific global approaches are tightly coupled to particular model families. Prototype methods often depend on deep encoders or latent representations \cite{ming2019interpretable,gee2019explaining,zhang2020tapnet,das2020interpreting,tang2020interpretable}, and shapelet-based approaches provide subsequence patterns but typically trade off interpretability with model- and task-specific design choices \cite{ye2009time,theissler2022explainable}. Feature-based conceptual frameworks \cite{kusters2020conceptual} offer another route, but are commonly evaluated through classification accuracy and are therefore less suited to general prediction.

In summary, prior work offers strong tools for \emph{local} time-series explanations and promising subset selection methods in other modalities, but there remains a need for a \emph{reliable, model-agnostic} way to construct global, time-series–aware summaries that guides the application of local explanations.

\section{Algorithmic Approach}\label{sec:algo_app}
We define a time-series summary \TSS{} as a concise collection of samples that highlight a model's key behaviors, focusing on selected segments or events while remaining manageable for human review. The primary challenge lies in determining which time-series segments should be included.

\begin{figure}[tb]
  \includegraphics[width=8.2cm ]{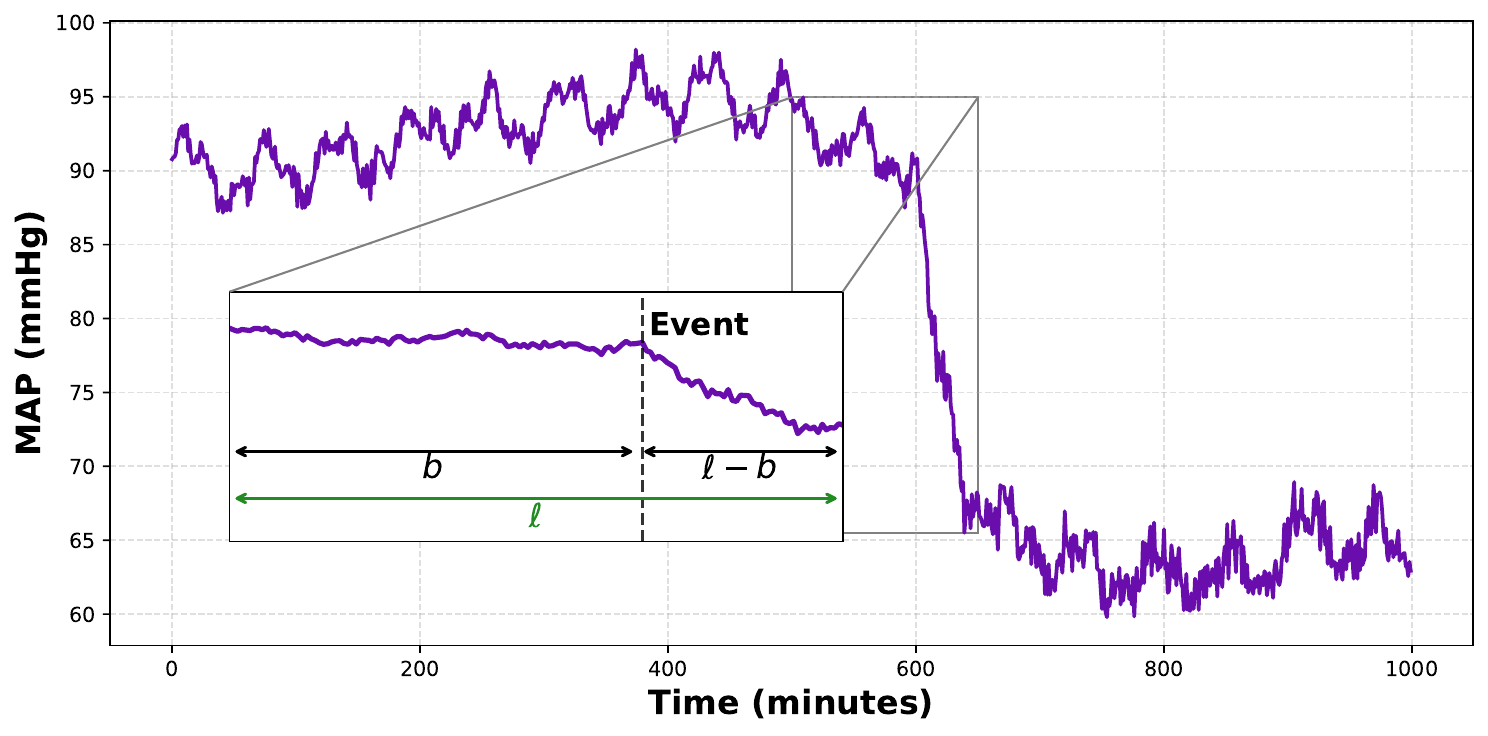} 
  \caption{Mean Atrial Pressure example: an event $e_t$ defines a window of length $l$, with $b$ time points before and $l-b$ after $t$.}
\label{fig:example} 
\end{figure} 

% \subsection{Running Example}
\paragraph{Running Example}\label{clinical}
Consider a clinician evaluating a blood pressure prediction model. She would not go over the entire dataset, opting instead for a carefully chosen set of illustrative examples. Predicting sudden drops is surely more important than predicting stable readings, yet importance extends beyond anomalies \cite{eini2022tell}; the relevance of $y_t$ depends on both clinical and temporal context.

\subsection{Notations}
\begin{figure}[htbp]
  \includegraphics[width=8.2cm ]{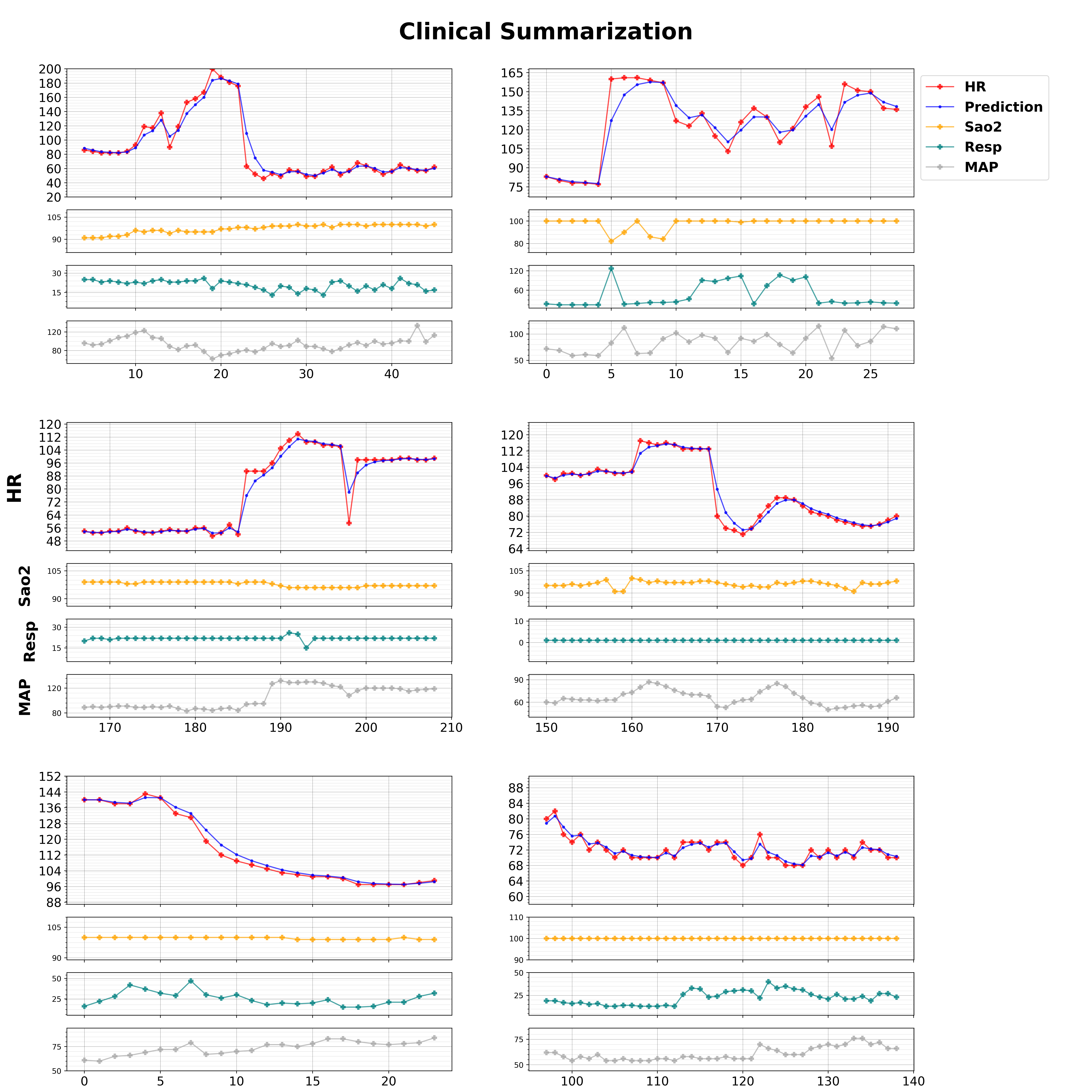}%{figs/vsg.pdf} 
  \caption{INSIGHTS-based \TSS{} of heart rate (HR) prediction task as displayed to the clinicians. The \TSS{} includes context information: SaO2, Resp, and MAP. The selected samples depict true events.}
\label{fig:user_INSIGHTS} 
\end{figure}

Time series datasets are constructed from multiple series. We denote a specific time point of series $i$ as $y_t^i$. Each series $i$ has $n_i$ time points, summing to $n$ time points in the entire dataset. For convenience, we will refer to a time point in any series simply as $y_t$ henceforth.
As mentioned in Section \ref{intro}, the significance of a specific time point $y_t$ is often derived from its temporal context. To account for this, we adopt a subsequence-based approach. For each time point $y_t$, we define a time series window $y_t \in e_t$ that includes $b$ time points before and $l-b$ time points after $t$, creating a time series snippet of length $l+1$ (see Figure~\ref{fig:user_INSIGHTS}).

\subsection{INSIGHTS Method}
The INSIGHTS algorithm enables utilizing of various \textit{utility functions} that reflect the significance of a time series sample from multiple perspectives. A utility function $f$ assigns an importance score for each time-point $y_t$ according to the aspect it represents. These scores are subsequently used to aggregate the samples into a comprehensive macro explanation. The primary criteria for selecting samples to comprise a \TSS{} are: (1) Importance, as indicated by utility scores; and (2) Diversity. The complete protocol for generating INSIGHTS-based \TSS{} is detailed in the following subsection, and given as pseudo-code in \Cref{alg:INSIGHTS}. Code and experiments are available in the supplementary. 

\subsubsection{Utility Functions}
Utility functions specify \emph{what the user considers important} in a time series segment; INSIGHTS uses them only to rank candidates for inclusion in the summary. Accordingly, utilities should reflect domain knowledge or other task-relevant notions of importance. In the clinical running example, for instance, domain norms help identify clinically unstable segments, and a gradual increase in blood pressure may indicate an evolving event.
% Since utility functions serve as importance measures aimed at identifying significant time series behaviors, they should be based on relevant domain knowledge or hold other significance for the prediction. 
% In example \ref{clinical}, 
In addition, accounting for domain norms may indicate which samples refer to patients with volatile medical condition; slowly increasing blood pressure suggests an evolving clinical event. 

We offer three general functions that capture utility from time series-specific properties concerning $y_t$ and its context within $e_t$.
% These functions are based on utility scores presented in Eini-Porat ~\shortcite{eini2024aiming} as evaluation metrics that reflect different aspects of time series prediction importance, complementary to RMSE. 
These defaults are adapted from the utility scores of Eini-Porat~\shortcite{eini2024aiming} (originally proposed as complementary evaluation metrics to RMSE). We adapt them to produce point-wise importance (utility) scores and allow for observing model successful predictions rather than just its failures per aspect. Optionally, utilities can operate on a user-defined feature map $\phi(e_t)$ rather than raw values. They may be user-defined feature map $\phi$ that transforms the local window $e_t$ into a representation, $\phi(e_t)\in\mathbb{R}$, capturing aspects of the signal that may be more relevant for importance (e.g., frequency-domain features or morphology descriptors). 

% Utilities can then be defined over $\phi(e_t)$ by replacing occurrences of $y_t$ (and slopes computed from $y_t$) with the corresponding feature values.

The formulations below present a default choice $\phi(e_t)=y_t$ (identity). Our functions address fundamental time series behaviors:

\begin{itemize}
\item Overall trend: \begin{equation}
    U_{t} = W_{h} \cdot \max\left(Y_{t}, 0\right)^2 + W_{\ell} \cdot \max\left(-Y_{t}, 0\right)^2
\end{equation}
Where $Y_{t}$ is the slope over window $e_t$ around time point $y_t$ (default: $\phi(e_t)=y_t$), and $W_{h}$, $W_{\ell}$ are hyper-parameters reflecting importance for either increasing or decreasing trends. 
\item Exceeding normal range:
 \begin{equation}
    U_{r} = \max\left(\sigma_{h}\left(y_{t}\right), 0\right) + \max\left(\sigma_{\ell}\left(y_{t}\right), 0\right)
\end{equation}

$U_{r}$ is comprised of two sigmoid functions $\sigma_{h}$ and $\sigma_{\ell}$ which capture the utility around each normal range threshold. Meaning, $y_t$ values become important once they exceed domain-relevant norm values. However, for extreme values, further deviations from the norm do not lead to an additional increase in importance.  

\item Sudden changes trend deviation:
 \begin{equation}
    U_{td} = \left(Y'_{t} - Y_{t}\right) ^2
\end{equation}
Where $Y'$ represents the expected trend, the naive slope over $b$ time points leading to time point $t$. 
\end{itemize}

% In principle, INSIGHTS algorithm enables users to use a subset of these functions or define different functions that receive a set of time series data and outputs importance per time point that reflect the significance of an a time series sample from multiple perspectives.
The selection and calibration of utility functions are domain and task-dependent. Exceeding normal range is important for the scenario in the clinical example as it has medical meaning; highly consistent values may indicate the patient is in pain, while low consistent values may reflect blood loss or infection. $U_r$ hyper parameters can reflect that. However, in stock price prediction, there is no natural normal range, making the $U_r$ function less relevant for this domain. Users may avoid selecting this function altogether. We note that the \textit{utility functions} above are primarily designed to capture time-series behavior which drives prediction tasks. For classification, utility functions may need to reflect behaviors that correspond with the task.

\subsubsection{Main Algorithm: Sample Pick}

INSIGHTS proceeds in rounds. Each round selects a utility perspective, proposes top-importance candidates, then adds the one that most increases behavioral diversity.

Here, we detail the proposed algorithm, as presented in \Cref{alg:INSIGHTS} for the construction of \TSS{} of size $m$.
We aim to construct a subset of subsequence samples $V$ that collectively offer informative insights. To this end, we address both the importance and diversity by calculating their importance using a utility functions set $\mathcal{F}$. The subset $V$ is iteratively expanded by incorporating the next sample with highest importance score that enhances diversity within the current set. In Example \ref{clinical}, if $V$ initially includes a segment showing elevated blood pressure, other important aspects, such as an increasing trend, may also be present. So, the algorithm prioritizes adding events with diverse behaviors (e.g sudden or decreasing trends).

\emph{Initialization} For every utility function $f \in \mathcal{F}$, utility scores per sample are calculated and assigned to the correspondent \emph{bucket of importance} $U_f$. Based on $U_f$, corresponding queue $q_f \in Q$ are then sorted internally (lines \ref{ln:loop_f} - \ref{ln:end_loop_f}). 
The order of buckets (line \ref{ln:init_bo}) is determined by the order of the maximum normalized utility scores: $rl_f(U_f) = \underset{i \in N}{\max}\frac{U_f^i}{std\left(U_f\right)}$. This procedure ensures that the sample with the highest utility score is selected.

\emph{Sample Selection} Next, $Q$ is iteratively traversed (lines \ref{ln:event_pick_loop} - \ref{ln:end_event_pick}). In each iteration, a set of  $m_c$ candidates $C$ with the highest score within are selected (line \ref{ln:select_candidates}). For each candidate $e_t \in C$ which satisfy overlap constraints with respect to $V$, diversity score reflecting information gain (line \ref{ln:diversity}).

The proposed \textit{diversity} computation relies on a straightforward approach for signal distance; a pairwise sample distance matrix $M_D$ based on dynamic time warping (DTW) \cite{tormene2009matching}, which aligns by adjusting for differences in timing or temporal scale. Next, $e_t \in C$ with maximal minimum distance from $e_t' \in V$ is selected: 

\begin{equation}
    \argmax_{e_t \in C} min_{(e,e_t'), e_t' \in V} (M_D(e_t,e_t')).
    \label{eq:timewrap}
\end{equation}

To evaluate the effect of a time series-specific diversity, we also consider an alternative calculation that does not explicitly account for temporal context. This alternative follows the \textit{critic} selection procedure  \cite{Kim2016ExamplesInterpretability}, where $V$ serves as the \textit{prototypes}: 

\begin{equation}
    \argmax_{e_t \in C} \left|\mathbb{E}\left(V\right) - \mathbb{E}\left(V \cup \left\{U(e_t)_f\right\}\right)\right|.
    \label{eq:mmd}
\end{equation}

We refer to these alternatives as \emph{INSIGHTS-TW} and \emph{INSIGHTS-critic}, respectively. 

Finally, $e_t$ which maximize diversity is added to $V$ (line \ref{ln:enter_examples}) and the non-overlapping events return to $q_f$ (line \ref{ln:add_back_candidates}). 
The algorithm repeats the above-mentioned steps until $m$ events are selected (lines \ref{ln:enter_examples} - \ref{ln:end_event_pick}). 
\emph{Prototypes} Selection of average sample - we allow the user to specify $m_p$ samples with median importance to include (line \ref{ln:add_proto}).
% Since the dataset is not solely characterized by most important samples, we allow the user to specify $m_p$ samples with median importance to include (line \ref{ln:add_proto}). They undergo similar verification (see Supplementary).

\paragraph{Time Complexity} Assuming utility functions set $\mathcal{F}$ can be computed in $O(n)$ time and memory, INSIGHTS is resource-efficient. DTW is computed only for small fixed-size subsequence windows $l \ll n$ and a fixed set of candidates $m_c \ll n$, so, the calculation of $M_D$ requires $O(m_c)$ iterations on fixed size,inputs. Making the sorting of utility buckets the dominant cost at $O\left(n\log\left(n\right)\right)$.

\begin{algorithm}[tb]
\DontPrintSemicolon
\SetKwComment{Comment}{$\triangleright$\ }{}

\SetKwInOut{Input}{input}\SetKwInOut{Output}{output}
\SetKw{KwBy}{by}
\SetAlgoLined
\Input{$Y, m, m_c, m_p, \mathcal{F}$ \Comment*[r]{Y is dataset}} 
\Output{$V$}
\SetKwComment{Comment}{$\triangleright$\ }{}
\BlankLine
    $Q \gets \left[\right]$ ;$ V \gets \emptyset$ ; $U \gets \left[\right]$ \label{ln:init_Q}\; 
    % $V \longleftarrow \emptyset $ \label{ln:init_V}\;
     % \label{ln:init_U}\;
    \ForEach{$f \in \mathcal{F}$\label{ln:loop_f}}{
        $U_f \longleftarrow f(Y)$ \label{ln:init_uf}\;
        $q_f \longleftarrow \underset{e \in Y}{\argsort}~U_f$ \label{ln:init_qf}\; 
    }\label{ln:end_loop_f}
    $BucketList \longleftarrow SetBucketOrder\left(U \right)$\label{ln:init_bo}\;

    \While{$\left|V\right| < m$ \label{ln:event_pick_loop}}{
        $i \longleftarrow BucketList.next()$ \label{ln:get_next_bucket_id}\;
        $q_f \longleftarrow Q\left[i\right]$\label{ln:get_bucket}\;
        \eIf{$V$ is empty\label{ln:check_first_pick}}{
            $y_t \longleftarrow q_f.pop()$\label{ln:get_first_item}\;
            $V \longleftarrow V \cup \left\{e_t\right\}$ \Comment*[r]{Extended window} \label{ln:enter_first}
        } 
        {\label{ln:start_other_picks}
            $C \longleftarrow selectCandidates\left(q_f, m_c\right)$\label{ln:select_candidates} \;
            $y_t \longleftarrow \underset{y_t \in C}{argmax}\left[DiversityScore\left(V,C\right)\right]$\label{ln:diversity}\;
            $addBackToQueue\left(C \setminus \left\{y_t\right\}, q_f \right)$ \label{ln:add_back_candidates}\;
            $V \longleftarrow V \cup \left\{e_t\right\}$ \Comment*[r]{Extended window}\label{ln:enter_examples}
        } \label{ln:end_other_pick}
    } \label{ln:end_event_pick}
    $V \longleftarrow appendPrototypes\left(V, Q, m_p, m_c, Y \right)$ \Comment*[r]{Append prototypes to $V$, ensure non-overlap} \label{ln:add_proto}
    \KwRet{$V$ \label{ln:return_v}}
 \caption{INSIGHTS algorithm}\label{alg:INSIGHTS}
\end{algorithm}

\section{Empirical Methodology}
We evaluate INSIGHTS through three complementary studies that target (i) computational performance under controlled conditions, (ii) domain-expert utility in a realistic clinical setting, and (iii) whether summaries measurably improve human understanding in a prediction task. Event capture tests whether INSIGHTS selects \TSS{} segments that are both important and diverse; interviews test whether summaries support expert sensemaking in a clinical workflow; and the user study tests whether summaries improve global understanding of model behavior in a prediction task. Together, these studies assess both the algorithmic properties of the produced \TSS{} (relevance and diversity) and their practical value for prospective users. 
The computational study examines whether the algorithm identifies important events and produces a more diverse \TSS{}. This is assessed through a quantitative comparison of its ability to capture pre-annotated events in two time-series datasets against three baseline methods. The two user-facing studies then evaluate the practical benefits of INSIGHTS \TSS{} for prospective users: semi-structured interviews with clinical domain stakeholders and a user study comparing the two leading summary approaches, where participants are presented with \TSS{} derived from stock data. To execute these evaluations, we compare our proposed approach to baselines based on existing model-agnostic methods that are applicable to prediction tasks, along with one naive random pick approach designed to isolate the effect of the \TSS{} itself. However, as previously discussed in \cref{background}, these methods -- MMD-critic \cite{Kim2016ExamplesInterpretability} and ProtoDash \cite{gurumoorthy2019efficient,aix360} -- do not inherently account for temporal context. To ensure a fair compression, we run these methods over the same candidate windows $e_t$. This modification allows them to implicitly use temporal context by considering $e_t$ as they would have considered image data, enabling an evaluation of their applicability to time-series tasks. 

\subsection{Computational Evaluation - Events Capture} \label{subsection::eve_capture}
The primary objective of this experiment is to compare the performance of INSIGHTS-TW and INSIGHTS-critic with baseline \TSS{} methods in accurately capturing time-series events while evaluating their overall diversity. To achieve this, datasets with pre-annotated events of various types were required. A secondary objective is to assess resource consumption, as baseline methods are known for their high computational demands. Performance is evaluated across multiple \TSS{} sizes appropriate for human review.

\textbf{Datasets} 2 datasets were tested: (1) synthetic time series dataset, as introduced in Eini-Porat~\shortcite{eini2024aiming}, comprises 1000 time series of a seasonal nature, each 500 time steps long and annotated events. The events are categorized into four types: evolving events, two kinds of sudden surges, and out-of-bounds events. (2) PhysioNet Sleep-EDF Expanded Database \cite{goldberger2000physiobank}, including 120 overnight EEG recordings. We extract clinically meaningful EEG events using standard signal-processing pipelines (see Supplementary), producing window-level annotations for micro-arousals, sleep spindles, K-complexes, and slow waves.

\textbf{Baselines} 
Three subset selection methods are evaluated: Random pick, MMD-Critic, and ProtoDash.

\textbf{Evaluation} Performance is measured by assessing \textit{event coverage}, proportion of event types captured by a given \TSS{} approach, and \textit{event examples}, proportion of $e_t$ which contain at least one event. These complementary measures reflect diversity and importance. For INSIGHTS variants, utilities are computed on domain-standard feature maps $\phi(e_t)$ (see Supplementary); in the EEG setting, $\phi$ is extracted per window using standard time--frequency / event-detection features, and we instantiate $U_r$, $U_t$, and $U_{td}$ accordingly. Additionally, regarding resource consumption, runtime and memory usage are evaluated.

\subsection{Expert Study}
In this study, semi-structured interviews were conducted with three intensive care (ICU) clinicians, representing prospective users. Since patient health is an evolving process, time series data is prevalent in healthcare, and as machine learning models are deployed, the need for explainability becomes critical. 
To anchor the discussion in a plausible scenario for evaluation, summaries by Random, INSIGHTS and the top performing baseline approach from experiment \ref{subsection::eve_capture}, were displayed. These summaries are comprised of examples from eICU dataset \cite{pollard2018eicu} and include display additional context relevant for ICU monitoring \cite{ivaturi2021comprehensive}: heart rate prediction model performance was shown alongside typical ICU vital signs such as heart rate (HR), blood oxygen saturation (SaO2), and respiration rate (Resp) for each example (see Supplementary).

To guide the interview, the clinicians were asked questions about performance and behavior of the heart rate prediction model (e.g., \textit{Based on these presented example summary, could you make observations with respect to model behavior?}), followed by broader inquiries regarding information conveyed by examples, and other requirements (e.g., \textit{Do time series example summaries provide a complete explanation?}). The display order of summaries was randomized to ensure questions about model behavior were first answered for a single summary, but broader conceptual questions refer to all summaries. The interviews aimed to explore potential requirements from time-series summaries in the clinical context. Open-ended questions were employed, transitioning from specific to general topics, to avoid introducing biases and allow user needs to emerge organically. Interview notes were summarized by the first author. In line with the questions, emerged concepts were sorted into topics referring to model behavior, examples and summary and applications.  

\subsection{User Study}

This study aims to assess the understanding of model prediction behaviors, as they are typically responsible prediction performance.
Initially, we aim to verify that human experts can deduce a time series prediction model's qualitative behavior based on example summaries. The primary goal is to determine whether an INSIGHTS-based summary enhances the understanding of model behavior. To achieve this, we generated summary explanations over stock data downloaded from  \cite{yahoo_finance} using the two leading methods identified in the previous study. The dataset includes stock prices for Microsoft, Amazon, and GE from October 1990 to October 2023 (or issue year). A naive regression model, which predicts stock values based on the previous observation and noise, was fitted. Participants were tasked with understanding this behavior based on the given example summaries.

\textbf{Participants, materials and recruitment} Participants were recruited through university announcements and social media advertisements, targeting students from STEM programs to simulate semi-expert users. Participants were compensated \$8 for a 15-minute experiment, with a \$20 bonus raffled among the best performers. Individuals who were colorblind or not proficient in interpreting graphs were excluded due to the nature of the experiment. The study was approved by our institutional review board.

% The experiment was conducted using the Qualtrics system. \beini{Do we need to say this?->} \ofra{no and you also don't need to mention qualtrics. but do say that the study was approved by the unviersity's institutional review board} We note that initial attempt to run the experiment with Prolific users aimed at reaching a larger sample was unsuccessful, as many users submitted responses in less than five minutes, skipped text questions, and overall performance appeared random across all groups.

% \paragraph{Protocol}
\textbf{Protocol} 
Understanding time series data requires graph proficiency, unlike image or text domains. Therefore, participants first received a graph proficiency introduction, explaining the plot and asking a basic question to confirm their understanding. 
Next, participants were presented with a six-example summary of the model predictions with respect to the actual stock trajectories and were asked \textit{How do you think the model works} to prompt thoughtful consideration of the summary. Subsequently, participants were presented with four randomly sampled questions depicting stock trajectories and three possible model predictions, from which they needed to identify the naive model they had seen. 
Finally, participants completed a short questionnaire based on \cite{hoffman2018metrics} aimed to measure subjective assessment of the summaries and explanation satisfaction.

\section{Results}

\begin{figure*}[tb]
\centering
\begin{subfigure}[tb]{0.3\linewidth}
\begin{minipage}[c]{0.08\linewidth}
\raggedleft \textbf{(a)}
\end{minipage}\hspace{0.5em}%
\begin{minipage}[c]{0.88\linewidth}
\includegraphics[width=\linewidth]{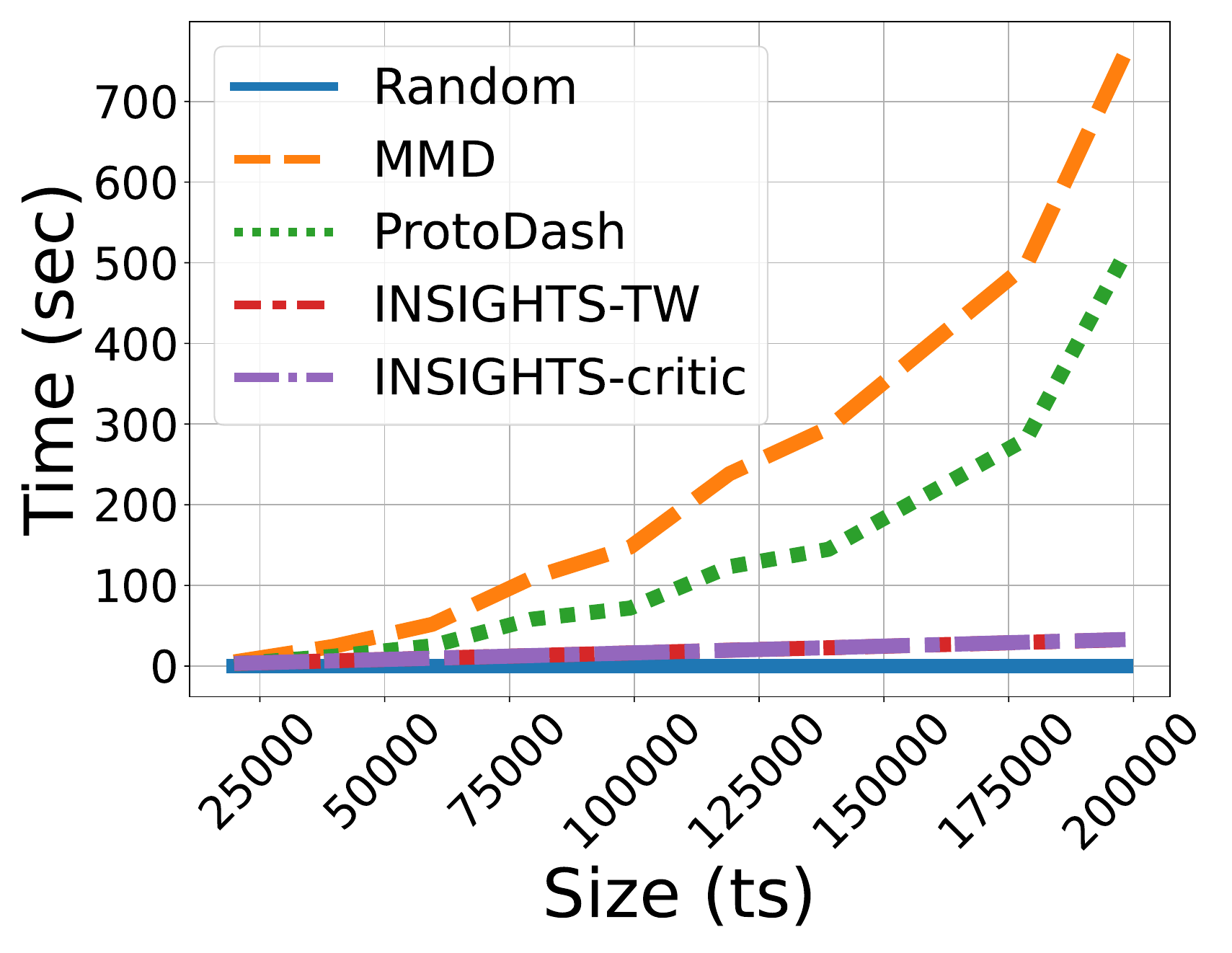}
\end{minipage}

\end{subfigure}
\begin{subfigure}[tb]{0.3\linewidth}
\begin{minipage}[c]{0.08\linewidth}
\raggedleft \textbf{(b)}
\end{minipage}\hspace{0.5em}%
\begin{minipage}[c]{0.88\linewidth}
\includegraphics[width=\linewidth]{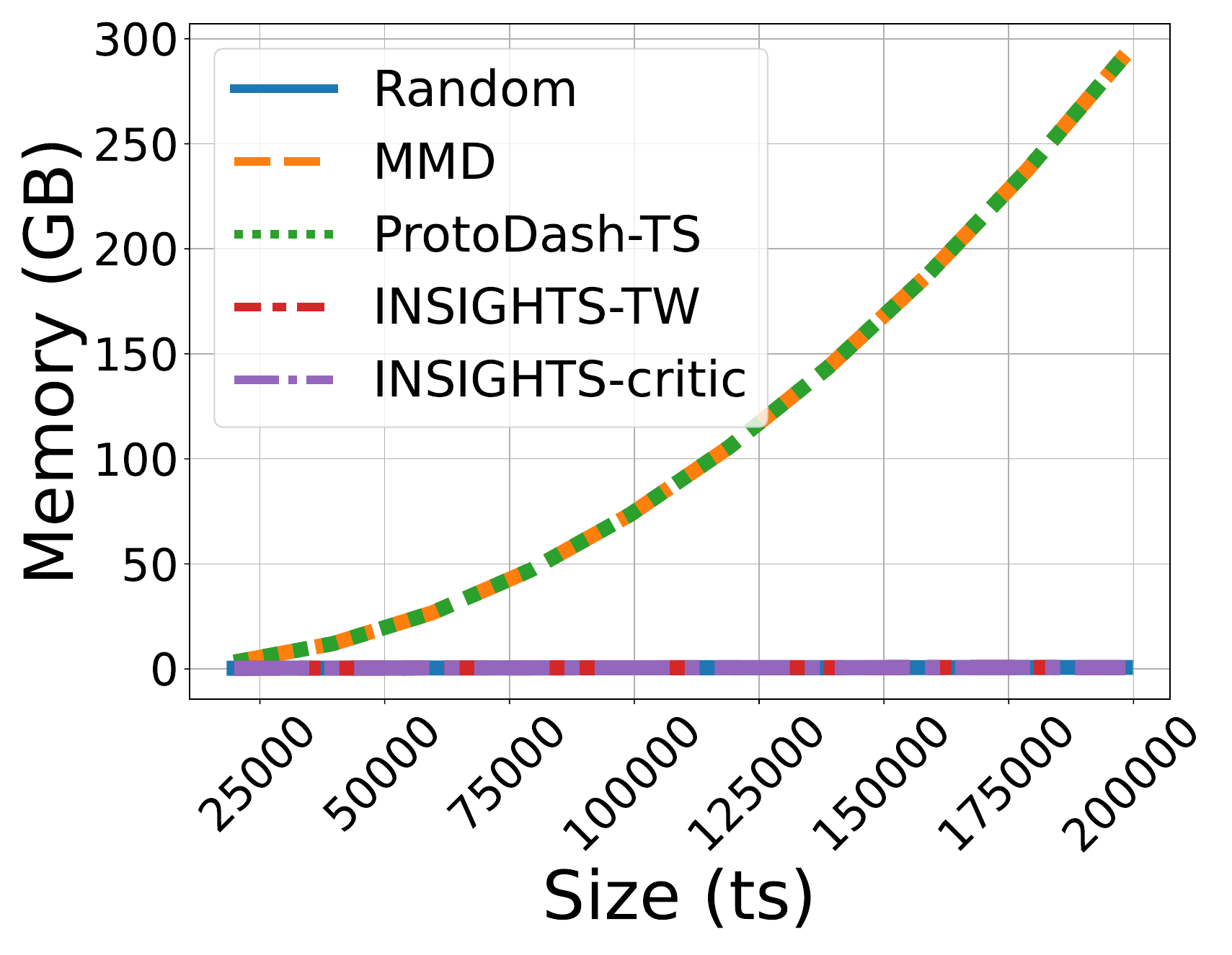}
\end{minipage}
\end{subfigure}
\caption{(a) Dataset size versus runtime in seconds. As the dataset size increases, ProtoDash and MMD exhibit significantly escalating runtimes, while INSIGHTS and Random methods maintain a moderately linear runtime.
(b) Dataset size versus memory usage. ProtoDash and MMD have memory usage significantly higher by compared to INSIGHTS and Random methods, which appear similar at this scale.}
\label{fig:memory}\label{fig:runtime_all}
\end{figure*}

\subsection{Events Capture} 
The results, summarized in Table \ref{table:quat_res}, indicate that the INSIGHTS-TW method achieves complete event-type coverage across all summary sizes. INSIGHTS-critic and ProtoDash both achieve 75\% coverage for smaller summaries, with ProtoDash reaching full coverage for sizes $\geq$ 10, while INSIGHTS-critic reaches full coverage for sizes $\geq$ 20. This difference is likely due to its non-time-series-specific diversity calculations. However, MMD-critic's coverage remains low, even compared to the Random approach. While INSIGHTS summaries slightly outperform ProtoDash in the percentage of event examples, both methods significantly outperform MMD-critic and Random, achieving near-perfect performance. Attempts to improve MMD-critic-based summaries by adjusting the ratio of prototypes to critics did not yield significant performance gains.

\begin{table}[htb]
\centering
\begin{tabular}{lcrr}
\toprule
\textbf{Size} & \textbf{Model} & \textbf{\begin{tabular}[c]{@{}c@{}}Event \\ Coverage (\%)\end{tabular}} & \textbf{\begin{tabular}[c]{@{}c@{}}Event \\ Examples (\%)\end{tabular}} \\
\midrule
% 4  & Random          & 25           & 25           \\
%                     & MMD             & 0            & 0            \\
%                     & ProtoDash       & 75           & \textbf{100} \\
%                     & INSIGHTS-TW     & \textbf{100} & \textbf{100} \\
%                     & INSIGHTS-critic & 75           & \textbf{100} \\
\midrule
6  & Random          & 25           & 16.67        \\
                    & MMD             & 0            & 0            \\
                    & ProtoDash       & 75           & \textbf{100} \\
                    & INSIGHTS-TW     & \textbf{100} & \textbf{100} \\
                    & INSIGHTS-critic & 75           & \textbf{100} \\
\midrule
8  & Random          & 25           & 12.5         \\
                    & MMD             & 50           & 25           \\
                    & ProtoDash       & 75           & \textbf{100} \\
                    & INSIGHTS-TW     & \textbf{100} & \textbf{100} \\
                    & INSIGHTS-critic & 75           & \textbf{100} \\
\midrule
10 & Random          & 25           & 12.5         \\
                    & MMD             & 50           & 36           \\
                    & ProtoDash       & \textbf{100} & \textbf{100} \\
                    & INSIGHTS-TW     & \textbf{100} & \textbf{100} \\
                    & INSIGHTS-critic & 75           & \textbf{100} \\
% \midrule
% 20 & Random          & 25 & 20 \\
%    & MMD             & 50 & 25 \\
%    & ProtoDash       & \textbf{100} & 90 \\
%    & INSIGHTS-TW     & \textbf{100} & \textbf{100} \\
%    & INSIGHTS-critic & \textbf{100} & \textbf{100} \\
\bottomrule
\end{tabular}
\caption{\TSS{} approaches performance for different \TSS{} sizes. 
INSIGHTS-TW approach exhibits maximal performance for all.}
\label{table:quat_res}
\end{table}

On the Sleep-EDF EEG dataset, INSIGHTS-TW achieves 80\% event-type coverage and 100\% event examples across all tested summary sizes, whereas Random achieves 0\% on both measures. In contrast, ProtoDash and MMD baselines could not be evaluated on this dataset: They exhausted system resources and failed to complete, despite utilizing a server equipped with an Intel Xeon Gold 6342 CPU (96 cores) and 500 GB RAM, even when restricted to 30\% of the recordings, questioning their ability to operate on large physiological time series datasets.% We analyze these resource costs in the runtime and memory experiments. %(Figures \ref{fig:runtime_all} and \ref{fig:memory}).

% Furthermore, although INSIGHTS-based summaries may achieve comparable performance to the ProtoDash baseline for certain summary sizes, their runtime is significantly shorter. 

This motivates our runtime analysis: even when summary quality can be comparable, INSIGHTS is substantially faster than ProtoDash. As illustrated in Figure \ref{fig:runtime_all}, when examining overall runtime versus dataset size, baseline summary approaches escalate quadratically, whereas INSIGHTS methods increase in a moderate, linear fashion. Moreover, increasing dataset size substantially impacts memory usage by ProtoDash and MMD methods.%, which hinders empirical runtime performance (see Figure \ref{fig:memory}). 

\subsection{Interview}
\textbf{Behavior Insights} Clinicians characterized the model as ``good'' and trustworthy across all sample summaries presented. 
% The most comprehensive insights into the model's behavior were derived from the INSIGHTS summary, followed by ProtoDash, and then Random.
INSIGHTS prompted comments on a broader set of behaviors—including oscillations, trend adaptation, sudden-event detection, and a possible recency bias—whereas ProtoDash mainly led to observations about conservativeness and trend handling, and Random rarely elicited any insights.
Notably, when the INSIGHTS summary was reviewed first, the clinician did not gain additional insights from the subsequent summaries. Conversely, when either ProtoDash or Random were reviewed first, clinicians identified significantly more insights upon subsequently reviewing the INSIGHTS summary. Across all summaries, the model was described as conservative, particularly in detecting sudden events and adapting to trends. Observations regarding the model's handling of oscillations were exclusively associated with the INSIGHTS summary. Furthermore, one clinician noted a bias in the model toward the most recent three steps, a finding that aligns with the model's underlying decay parameter.

\textbf{Summary Quality and Example Relevancy} Clinicians defined an informative sample as one containing significant events such as evolving or changing trends. One clinician mentioned that presenting an example with no events could be informative as it reflect model behavior routine observations. In the INSIGHTS-based summary, all but one example were considered informative, each offering at least one insight into the model's behavior. In contrast, the Random summary had only one or two informative examples, with the rest deemed useless. Although most examples in the ProtoDash summary were individually useful, they lacked diversity, leading to overlapping examples.

\textbf{Completeness}  Although clinicians understood the model's overall behavior, they found it challenging to fully grasp its decisions without additional information relevant to specific predictions. This includes feature importance measures, ``importance heatmap'' or model weights relevant to a prediction.

\textbf{Applications} According to the clinicians, these summaries can serve as an introduction to the model, helping to establish and calibrate trust through interpretable examples. They also reveal the model's strengths and weaknesses by highlighting instances where it performs well or poorly, which can be valuable for model selection. One clinician specifically noted the usefulness of these summaries in model selection. Additionally, a practical use case emerged in two interviews: utilizing these summaries during staff shift changes to present patient highlights. This application can showcase significant instances in patient vital signs from the previous shift, enhancing continuity of care and ensuring effective communication of critical information between healthcare providers.

\subsection{User Study}
\begin{figure}[tb]
\centering
\includegraphics[width=0.7\linewidth]{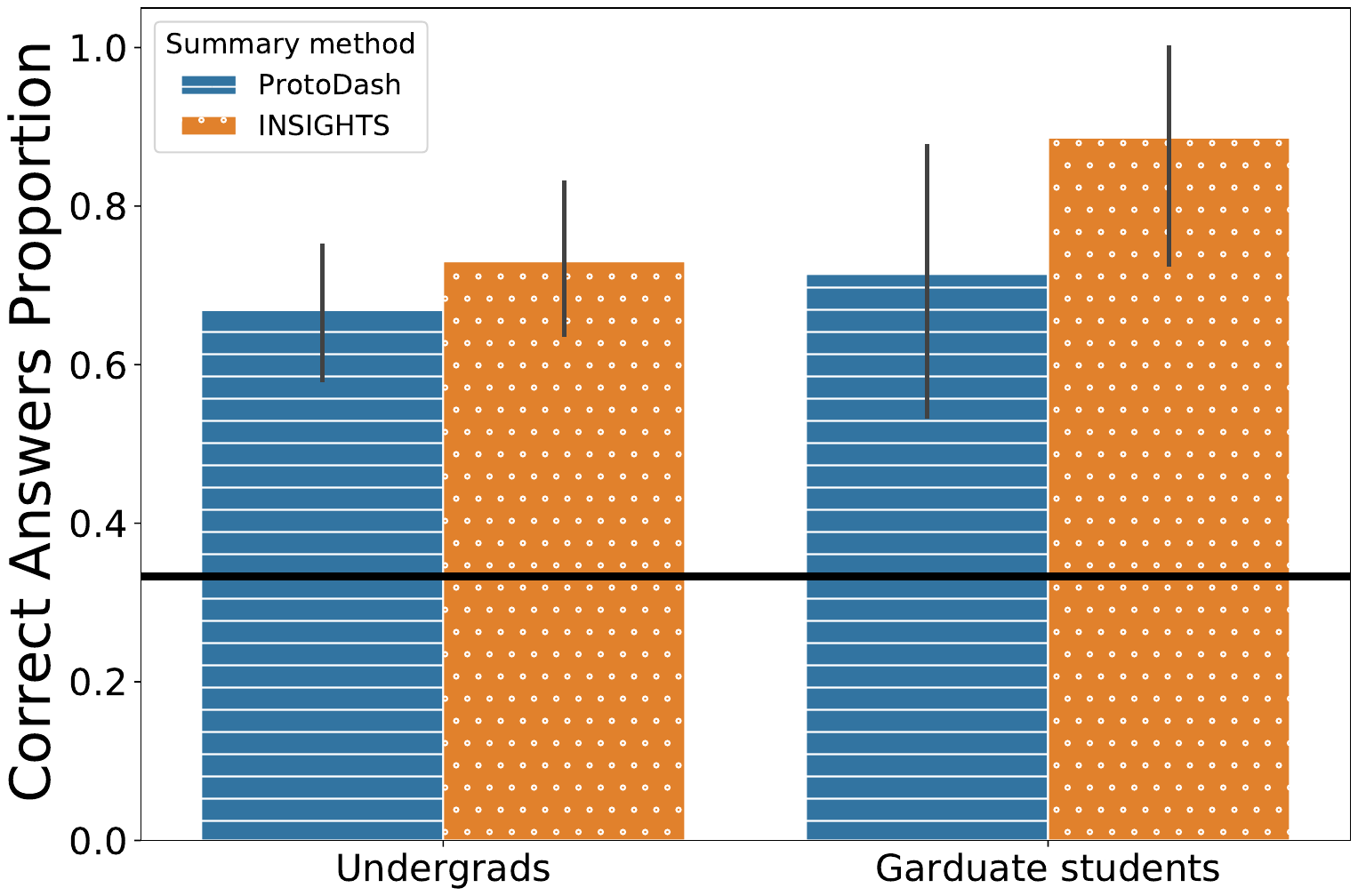}
\caption{Percentage of correct answers per \TSS{} type. Performance is illustrated for graduate and undergraduate students. The horizontal line is random choice performance.}
\label{fig:user}
\end{figure}

A total of 103 (42\% female) participants took part in this study, with 54 receiving a ProtoDash-based summary and 49 receiving an INSIGHTS-based summary. Participants in both groups seemed to identify the correct model behavior significantly better than a naive choice $0.33$ ($pvalue \leq 0.01$). The INSIGHTS summary group had a slightly higher performance than the ProtoDash group, 0.77 (95\% CI: 0.71, 0.83) versus 0.68  (95\% CI: 0.62, 0.74), though the difference was only marginally statistically significant ($pvalue = 0.08$). Given that graduate students performed better in this task, we report results within these two sub-groups in \Cref{fig:user}. Notably, while two questions had a high initial success rate of approximately 80\%, the other two revealed a 20\% improvement for the INSIGHTS summary group (see examples in Supplementary). 
Regarding trust and satisfaction questionnaires, participants who examined INSIGHTS-generated summaries rated the explanations as more complete ($pvalue = 0.03$) and the task as less demanding ($pvalue = 0.04$). Overall, they rated the summary as more reflective of model reliability compared to the ProtoDash group and reported a lower cognitive workload. However, the differences between the two approaches in the rest of questionnaire were nonsignificant.

\section{Discussion} \label{discussion}

INSIGHTS is a user-centric approach to evaluate time series datasets using domain- and context-specific importance. Its transparency and configurability distinguish it from time series-specific interpretability methods \cite{jain2019attention,theissler2022explainable}. INSIGHTS performs well in generating \TSS{} in terms of both diversity on the synthetic and Sleep-EDF datasets, and captures salient events when the chosen utility functions align with the domain (see Supplementary). While ProtoDash selects informative samples, it yields less diverse subsets in both the quantitative study and the domain-expert evaluation. This likely stems from diversity calculations that are not time-series aware; for example, a sharp surge at time point $t$ may be treated as different from the same surge at $t{+}1$. Similarly, INSIGHTS-critic, which uses non-time-series-specific diversity, struggles to capture diverse events for small \TSS{} sizes. The EEG experiment points to a practical constraint: ProtoDash could not be evaluated on Sleep-EDF due to resource exhaustion, underscoring feasibility limits for these baselines at real-world scale.

In the user study, self-reports suggest that INSIGHTS-based \TSS{} improves satisfaction, trust, and perceived cognitive load. The main task objectively measures understanding of a prediction pattern and therefore benefits from summaries containing high-utility samples. While the overall improvement was not statistically significant, harder questions—those requiring deeper understanding—were better supported by the more diverse INSIGHTS-based \TSS{}. Graduate students, whose experience better matches the intended user profile, benefited more from INSIGHTS-based \TSS{}.

% Overall, INSIGHTS achieves better or comparable results to ProtoDash-based \TSS{} while being substantially more resource-efficient. ProtoDash's complexity is polynomial in both \TSS{} size and input size \cite{gurumoorthy2019efficient}, whereas INSIGHTS is log-linear in input size and linear in \TSS{} size. This makes INSIGHTS a practical front end for applying model-agnostic local explanations to selected samples. Its efficiency and configurability also support querying large time-series repositories even without a predictive model, as suggested in the interviews. Interviewees further preferred additional context—such as feature local explanations—alongside selected samples to make summaries more actionable.

Overall, INSIGHTS achieves better or comparable results to ProtoDash-based \TSS{} while being more resource-efficient. ProtoDash's complexity is polynomial in both \TSS{} size and input size \cite{gurumoorthy2019efficient}, whereas INSIGHTS is log-linear in input size and linear in \TSS{} size. This makes INSIGHTS a practical front end for applying model-agnostic local explanations to selected samples. This way, the global summary provides a principled set of segments on which local, model-agnostic explainers can be applied. Its efficiency and configurability also support querying large time-series repositories without a predictive model, as suggested in the interviews. Interviewees further preferred additional context—such as feature local explanations—alongside selected samples to make summaries more actionable.

A primary limitation is the lack of additional public real-world time-series datasets with explicit event annotations for repeating the event-capture evaluation; many benchmarks instead provide classification labels (e.g., anomaly or arrhythmia detection), which are less suitable for evaluating event capture in prediction settings. Since event-level annotations for forecasting behaviors are rare (beyond task labels), synthetic data provides controlled ground truth for event capture; we complement it with a real-world event-capture study on Sleep-EDF and two user-facing evaluations on real-world datasets (ICU and stock data). To counter that, additional evaluations with users were conducted including additional two real-world datasets. INSIGHTS also uses greedy selection, which does not guarantee optimality but improves efficiency. Finally, the utilities presented are designed for univariate series; a naive multivariate adaptation treats each channel independently and ignores interactions. Multivariate extensions—via multivariate utilities or feature maps that capture interactions—are natural next steps within the same framework and time-warp-based diversity.
\bibliographystyle{named}
\bibliography{ijcai26}

\end{document}